# A Formal Tool for Verification of Probabilistic Spiking Neural Networks Based on Quotient Abstractions

Nikan Zandian Jazi[1], Elisabetta De Maria[1] and Christopher Leturc[1]

Université Côte d'Azur, CNRS, I3S, France

**Abstract.** Spiking Neural Networks (SNNs) model biological neural dynamics more faithfully than classical artificial networks, but their stochastic, event-driven computation—rooted in ion-channel noise and unreliable synaptic vesicle release—demands probabilistic models for which deterministic abstractions are mathematically inadequate. Formal verification of such models via probabilistic model checking faces a fundamental barrier: the *state space explosion problem*, where the Discrete-Time Markov Chain (DTMC) encoding grows exponentially with the number of neurons. General-purpose quotient model abstractions [1] can in principle mitigate this growth by partitioning membrane potentials into equivalence classes, but a naïve application to SNNs discards synaptic weight information, limiting the properties that can be verified. This paper introduces a *weight-discretized quotient model abstraction* that maps continuous synaptic weights to a compact integer range while preserving the relative contribution of each synapse, and presents CogSpike, a unified workbench that integrates SNN design, simulation, and PRISM-based formal verification within a single isomorphic tool chain. The discretization is accompanied by formal correctness guarantees: a two-sided fidelity theorem confines any firing disagreement to a bounded gray zone around threshold, and an Asymptotic Silence theorem gives the exact limit guarantee that unforced neurons fall permanently silent. A topology-dependent scaling analysis shows that the state space reduction compounds exponentially—approximately $17\times$ per neuron for discretization parameter $W = 3$—enabling verification of networks that are otherwise intractable, as confirmed empirically across seven canonical topologies.

## 1 Introduction

Spiking Neural Networks (SNNs)—the third generation of artificial neural networks [2] —are modelled as directed graphs whose nodes represent neurons and whose edges represent synaptic connections that can be either *excitatory* (positive weight) or *inhibitory* (negative weight). Unlike rate-coded deep networks, SNNs communicate via discrete, asynchronous spikes whose precise timing encodes information alongside aggregate firing rates, making them a natural candidate for studying how real neural circuits process and transmit information. Among SNN neuron models, the Leaky Integrate-and-Fire (LIF) formulation provides analytical tractability, while biophysically detailed models such as Hodgkin–Huxley [3] and computationally efficient alternatives such as Izhikevich neurons [4] reproduce a wider repertoire of biological firing patterns at higher computational cost.

Understanding how the brain computes requires studying the *temporal dynamics* of neural circuits—how spikes propagate, interact, and give rise to emergent behaviours in small but functionally relevant network topologies such as chains, convergent motifs, and recurrent loops. However, modelling these dynamics demands a delicate balance. On one hand, biological neurons are inherently stochastic: ion-channel noise, unreliable synaptic vesicle release, and variable axonal delays introduce randomness at every stage of signal transmission [3, 5], making probabilistic models mathematically necessary. One promising avenue encodes the network as a Discrete-Time Markov Chain (DTMC) and applies probabilistic model checking, but this faces a fundamental barrier: the *state space explosion problem*, where the DTMC state space grows exponentially with network size, rendering verification intractable beyond a handful of neurons. A model that is biologically faithful yet computationally tractable enough for formal analysis represents a Pareto compromise between these two desiderata.

This paper pursues precisely that compromise. We propose CogSpike, a unified tool for probabilistic spiking neural networks that integrates three tightly coupled capabilities within a single isomorphic framework: (i) *simulation* of LIF-based SNN dynamics, (ii) *formal modelling* of the same networks as DTMCs for the PRISM model checker [6] (a tool that computes the probability with which a temporal-logic property holds over a DTMC by exhaustive numerical analysis, up to floating-point precision), and (iii) *automated model checking* of behavioural properties expressed in Probabilistic Computation Tree Logic (PCTL). The underlying neuron model employs a weight-discretized quotient abstraction that overcomes the limitations of naïve quotient models [1], which discard synaptic weight information when partitioning states into equivalence classes.

Concretely, the contributions are fourfold:

1. A **weight discretization scheme** that maps continuous synaptic weights to a finite discrete range while preserving threshold feasibility and relative synaptic contributions (Section 4.1).
2. **Formal correctness proofs**: a two-sided *discretization-fidelity* theorem that confines any firing disagreement to a bounded gray zone (with a constructive rule for choosing $W$), and an *Asymptotic Silence* theorem giving the exact limit guarantee that unforced neurons fall silent (Section 4.4).
3. A **topology-dependent state space analysis** with closed-form formulas for DTMC size as a function of network structure, validated empirically across seven canonical topologies (Section 6).
4. **CogSpike**[1], a unified workbench integrating SNN design, simulation, and formal verification, whose code generator produces a PRISM representation isomorphic to the simulation engine, enabling automated formal modelling and model checking (Section 5).

## 2 Related Work

### 2.1 Stochastic Approaches to SNN Modelling

Biological neurons exhibit significant trial-to-trial variability even under identical stimulation [7]. Classical approaches capture this stochasticity through three mechanisms: *escape noise*, which introduces a probabilistic firing threshold; *diffusive noise*, modelling stochastic spike arrivals via synaptic bombardment; and *slow noise*, which adds fluctuations to neuronal parameters [7]. These formulations underpin large-scale analyses of noisy integrate-and-fire networks [8], but they operate in continuous state spaces and do not yield discrete-state models amenable to exhaustive formal verification.

Among existing simulators, Brian 2 [9] supports stochastic firing thresholds via escape noise (SDE integration), while Nengo [10], NEST [11], and BindsNET [12] are limited to noise injection at the input level. Crucially, none of these platforms support probabilistic model checking.

Our model takes a different route: rather than adding continuous noise to a differential equation, we discretize the membrane potential into threshold levels and assign each level an explicit firing probability, yielding a finite-state probabilistic model that maps directly onto a DTMC. This enables exhaustive formal verification via model checking—a capability that is fundamentally unavailable with continuous-noise formulations.

[1]All code and experiments are available at https://github.com/QuietRocket/CogSpike.

### 2.2 Formal Verification of SNNs

De Maria et al. [13] pioneered the modelling of SNNs as timed automata, formalizing Leaky Integrate-and-Fire (LIF) neurons with parameter learning. Their work established key biological properties—tonic spiking, integrator behaviour, excitability—as formal verification targets, and introduced the Advice Back-Propagation (ABP) algorithm for supervised parameter inference driven by model-checking counter-examples rather than continuous gradients. This approach was extended to *neuronal archetypes*—primitive micro-circuits such as contralateral inhibition and parallel composition—allowing macroscopic network properties to be composed from formally verified building blocks [14].

More recently, Yao et al. [15] introduced the Refractory-evolve Probabilistic LI&F (RP-LI&F) neuron model, unifying discrete-time refractory dynamics with probabilistic spike generation. Their contract-based verification approach translates SNN topologies into Discrete-Time Markov Chains (DTMCs) and specifies behavioural properties using Probabilistic Computation Tree Logic (PCTL), enabling rigorous assume/guarantee contracts.

These approaches establish the feasibility of formal SNN verification but share a common limitation: the *state space explosion problem*. As network size grows, the DTMC state space grows exponentially, and general-purpose quotient abstractions [1] would lose synaptic weight information if applied naïvely. Moreover, no existing tool unifies probabilistic SNN simulation and formal verification. The present work addresses all three limitations.

## 3 Preliminaries

### 3.1 Spiking Neural Network Model

An SNN is modelled as a directed graph $G = (V, E)$, where directed edges represent unidirectional synaptic connections and $V = V_{\text{in}} \cup V_{\text{proc}} \cup V_{\text{out}}$ partitions into input, processing, and output neurons, and $E \subseteq V \times V$ represents directed synaptic connections with integer weights $w_e \in [-100, 100]$.

Each processing or output neuron $n \in V_{\text{proc}} \cup V_{\text{out}}$ follows Leaky Integrate-and-Fire (LIF) dynamics [3, 13]; input neurons in $V_{\text{in}}$ act as exogenous spike sources. Let $r \in [0, 1]$ be the leak factor and $T$ be the firing threshold. At each discrete time step $t$, the membrane potential integrates incoming weighted spikes and decays toward rest:

$$p_n(t+1) = \max\left(0, r \cdot p_n(t) + \sum_{i \in \mathrm{In}(n)} w_{i,n} \cdot y_i(t)\right) \quad (1)$$

where $y_i(t)$ is the spike event of presynaptic neuron $i$ at time $t$. When $p_n(t+1) \geq T$, neuron $n$ emits a spike ($y_n(t+1) = 1$) and resets to zero.

Firing is *probabilistic*: the potential maps to a discrete integer threshold level $L \in \{0, ..., k-1\}$ where $k$ is configurable (1–10), and each level yields a firing probability. Optionally, neurons implement a three-state refractory machine: Normal ($s = 0$), Absolute Refractory Period (ARP, $s = 1$), and Relative Refractory Period (RRP, $s = 2$), with reduced firing probability during RRP scaled by factor $\alpha$.

### 3.2 Model Checking and Temporal Logics

**Definition 1**. *A* Discrete-Time Markov Chain *(DTMC) is a tuple $\mathcal{D} = (S, s_0, \boldsymbol{P})$ where $S$ is a finite set of states, $s_0 \in S$ is the initial state, and $\boldsymbol{P} : S \times S \to [0,1]$ is the transition probability matrix satisfying $\sum_{s' \in S} \boldsymbol{P}(s, s') = 1$ for all $s \in S$ [1].*

Behavioural properties over state-transition systems are expressed using *Computation Tree Logic* (CTL) [1], a branching-time temporal logic built from four path operators: $\boldsymbol{X}\varphi$ (*neXt*—$\varphi$ holds in the immediate successor state), $\varphi_1 \boldsymbol{U} \varphi_2$ (*Until*—$\varphi_1$ holds along a path until $\varphi_2$ becomes true), $\boldsymbol{F}\varphi$ (*Finally*—$\varphi$ eventually holds, syntactic sugar for $\top \boldsymbol{U} \varphi$), and $\boldsymbol{G}\varphi$ (*Globally*—$\varphi$ holds at every state along the path).

*Probabilistic Computation Tree Logic* (PCTL) extends CTL to stochastic systems by replacing the universal/existential path quantifiers with a probabilistic operator $P_{\bowtie p}[\psi]$, which asserts that the probability of satisfying path formula $\psi$ meets the bound $\bowtie p$ [1, 16]. For instance, $P_{\geq 1}[\boldsymbol{F}(y_n = 1)]$ asserts that neuron $n$ fires with probability one, while $P_{\geq 1}[\boldsymbol{G}(y_n = 0)]$ asserts permanent silence.

The role of a *probabilistic model checker* is to compute, given a DTMC $\mathcal{D}$ and a PCTL property $\varphi$, the probability with which $\varphi$ is satisfied from the initial state by exhaustive numerical analysis, up to floating-point precision. In the context of SNN verification, this serves two purposes: (i) *validating model correctness*—confirming that the formal DTMC encoding faithfully reproduces expected neural behaviours (e.g., tonic spiking under sustained input, silence without input); and (ii) *studying the temporal dynamics* of small but functionally relevant neural configurations, such as quantifying spike propagation probabilities across chains or characterizing inhibitory gating in convergent motifs.

PRISM [6] is a probabilistic model checker supporting DTMCs. Models are specified as parallel compositions of *modules* with local integer variables and guarded probabilistic transitions, synchronizing via shared labels. The global state space is the Cartesian product of all module state spaces. PRISM offers an *explicit* engine (enumerates reachable states) and a *symbolic BDD* engine (uses Binary Decision Diagrams via the CUDD library), whose efficiency depends on variable ordering.

### 3.3 Quotient Model Abstraction

Quotient model abstraction [1, 17] reduces the DTMC state space by partitioning *probabilistically bisimilar* states—those yielding identical firing probabilities and, under every input, transitioning to equivalent successor classes—into equivalence classes. The resulting quotient is the coarsest PCTL-preserving partition, reducing the per-neuron state space from $|P_{\max} - P_{\min} + 1|$ values to $k+1$ classes (where $k$ is the number of threshold levels). However, a naïve application to SNNs treats all synapses uniformly, collapsing weight information: it cannot distinguish strong excitatory from weak excitatory or inhibitory contributions. The weight discretization scheme in Section 4 addresses this limitation.

## 4 Weight-Discretized Quotient Abstraction

The quotient model of Section 3.3 abstracts membrane potentials into equivalence classes but treats all synapses uniformly. This section introduces a *weight discretization scheme* that resolves this limitation while preserving the relative contribution of each synapse.

### 4.1 Weight Discretization Function

**Definition 2**. *Given a weight range* $[-w_{\max}, w_{\max}]$ *(typically* $w_{\max} = 100$*) and a discretization parameter* $W \in \mathbb{N}^+$*, the* weight discretization function $\delta_W : \mathbb{R} \to \mathbb{Z}$ *is:*

$$\delta_W(w) = \mathrm{round}\left(w \cdot \frac{W}{w_{\max}}\right) = \left\lfloor w \cdot \frac{W}{w_{\max}} \right\rceil \tag{2}$$

*mapping original weights to the discrete range* $[-W, W] \subset \mathbb{Z}$*.*

The function preserves relative weight magnitudes: for $W = 3$, strong excitatory ($w = 100$) maps to $\delta_3(100) = 3$, medium ($w = 67$) to $\delta_3(67) = 2$, weak ($w = 33$) to $\delta_3(33) = 1$, and inhibitory ($w = -50$) to $\delta_3(-50) = -2$. Weights are stored as static integer constants in the generated PRISM model, contributing no additional state variables.

The *weighted contribution* for neuron $n$ with discretized incoming weights $\{w_1^d, ..., w_m^d\}$ replaces the binary class evolution of the original quotient model:

$$C_n = \sum_{i=1}^{m} w_i^d \cdot y_i \tag{3}$$

where $y_i \in \{0, 1\}$ is the spike output of presynaptic neuron $i$ and $m$ is the fan-in.

### 4.2 Threshold Calibration

**Definition 3**. *The* discretized threshold *for a neuron with original threshold $T$ is:*

$$T_d = \left\lceil T \cdot \frac{W}{w_{\max}} \right\rceil \tag{4}$$

The use of the ceiling function ceil (rather than rounding) ensures $T_d \geq T \cdot W/w_{\max}$, so the discretized neuron is *at least as hard* to fire as the original. This conservative calibration underlies the Asymptotic Silence guarantee (Theorem 2).

### 4.3 Multiplicative Leak

In the discretized model, the membrane potential decays via the same multiplicative leak as the original LIF dynamics (Equation 1): the discretized update rule is

$$p'_n = \lfloor r \cdot p_n \rfloor + C_n \tag{5}$$

where $C_n$ is the weighted contribution from Equation 3 and $r \in [0, 1]$ is the leak factor. Since $r < 1$, the floor operation guarantees that $\lfloor r \cdot p_n \rfloor < p_n$ for all $p_n > 0$, ensuring strict decay in the absence of input. This preserves strict isomorphism between the simulation engine and the PRISM model.

### 4.4 Formal Proofs

This subsection presents the complete proofs of the key correctness properties.

**Two-Sided Discretization Fidelity.**

The weight discretization is a bounded-error quantization. Each $\delta_{W(w_i)} = \lfloor w_i \cdot W/w_{\max} \rceil$ lies within 1/2 of $w_i \cdot W/w_{\max}$, so for any pattern $\boldsymbol{y}$ with weighted sum $S = \sum_i w_i y_i$ and $m^* = \sum_i y_i$ active inputs the discretized sum $S_d = \sum_i \delta_{W(w_i)} y_i$ satisfies

$$S \cdot W/w_{\max} - m^*/2 \leq S_d \leq S \cdot W/w_{\max} + m^*/2. \tag{6}$$

The next theorem turns this error bound into a two-sided firing guarantee.

**Theorem 1**. *Let $\mathcal{N}$ be a neuron with threshold $T$ and fan-in $m$, and let $T_d = \lceil T \cdot W/w_{\max} \rceil$. For an input pattern with weighted sum $S$:*

1. (Completeness) *if $S - T \geq w_{\max}(m/2+1)/W$, the discretized neuron fires, i.e. $S_d \geq T_d$;*
2. (Soundness) *if $T - S > w_{\max} m/(2W)$, the discretized neuron does not fire, i.e. $S_d < T_d$.*

*Proof.* The ceiling gives $T \cdot W/w_{\max} \leq T_d \leq T \cdot W/w_{\max} + 1$. *Completeness:* by Equation 6 and $m^* \leq m$, $S_d \geq SW/w_{\max} - m/2 \geq TW/w_{\max} + 1 \geq T_d$, where the middle inequality uses $S - T \geq w_{\max(m/2+1)}/W$. *Soundness:* by Equation 6, $S_d \leq SW/w_{\max} + m/2 < TW/w_{\max} \leq T_d$, where the strict inequality uses $T - S > w_{\max} m/(2W)$. □

Equivalently, the discretized decision matches the original outside a *gray zone* of half-width $w_{\max} m/(2W)$ around threshold (with an extra $w_{\max}/W$ on the firing side from the ceiling), and the ceiling calibration biases borderline cases toward silence. The bound is constructive: to preserve a single-step firing with margin $\gamma = S - T > 0$, choose $W \geq w_{\max}(m/2+1)/\gamma$; a smaller $W$ widens the gray zone—the price paid for the state-space compression.

**Asymptotic Silence.**

The two-sided bound leaves a gray zone in which a single step may flip, but the multiplicative leak removes the ambiguity in the limit: once the input ceases, a subthreshold neuron is silent forever—an exact guarantee with no gray zone.

**Theorem 2**. *Let $\mathcal{N}'$ be a discretized neuron with potential $p_t < T_d$ and leak factor $r < 1$. If the input is zero for all $t' \geq t$ (i.e., $C_n = 0$ henceforth), then $\mathcal{N}'$ will never fire.*

*Proof.* Without input, $p_{t+1} = \lfloor r \cdot p_t \rfloor$. Since $r < 1$ and $p_t > 0$, we have $\lfloor r \cdot p_t \rfloor < p_t$; hence the potential is strictly decreasing and converges to the absorbing state $p = 0$. Since the trajectory is non-increasing and starts below $T_d$, the firing condition $p \geq T_d$ is never met. At $p = 0$, the threshold level is $L = 0$, which maps to firing probability zero, ensuring permanent silence. $\square$

**Remark (the gray zone is intrinsic).** No $W$ in the compression regime closes the gray zone. Exact single-step completeness needs $T_d \leq TW/w_{\max} - m/2$ and exact single-step soundness needs $T_d > (T-1)W/w_{\max} + m/2$; together these force $W > m \cdot w_{\max}$, which annihilates the reduction. Both failure modes are real at $W = 3$, $w_{\max} = 100$: a single weight $w = 80 = T$ rounds to $\delta_3(80) = 2 < T_d = 3$ (a fireable neuron silenced), while $w_1 = w_2 = 50 < T = 120$ rounds to $\delta_3(50) = 2$ each, summing to $4 = T_d$ (a subthreshold neuron made to fire). The absorbing silence of Theorem 2 is therefore the operative soundness guarantee —it is exact and matches the limit behaviour the verified PCTL properties (e.g. $\boldsymbol{FG}(y_n = 0)$ for losing neurons) actually require.

**Biological Property Preservation.**

The discretized model preserves the core LIF properties formalized by De Maria et al. [13]:

- **Tonic spiking.** Under constant input $C_{\text{in}}$, the neuron has non-zero firing probability iff the net gain per step overcomes the multiplicative decay, i.e., $C_{\text{in}} > T_d \cdot (1 - r)$. When satisfied, the potential accumulates toward the steady state $p_{\text{ss}} = C_{\text{in}}/(1 - r)$, yielding periodic probabilistic spiking.
- **Integrator.** The probability of immediate firing on simultaneous inputs reaches 1.0 iff $\sum \delta_{W(w_i)} \geq T_d$. Below threshold, the response is graded via the threshold level mapping.
- **Excitability.** The expected inter-spike interval decreases monotonically as input strength increases, since stronger input yields higher net accumulation per step and a faster approach to threshold.

## 5 The CogSpike Workbench

Building on the abstraction of Section 4, CogSpike turns the design–simulate–verify loop into a single tool. Whereas the simulators surveyed in Section 2 provide stochastic dynamics but no path to formal verification, CogSpike unifies *probabilistic* SNN *design*, *simulation*, and *verification* in a single desktop workbench. The tool is implemented in Rust with an immediate-mode GUI (egui), and its core design principle is **strict isomorphism**: the PRISM code generator produces a DTMC representation that is isomorphic to the simulation engine—both share the same mathematical model, namely the LIF dynamics of Equation 1, the three-state refractory machine, and the probabilistic firing logic —so that verification results faithfully analyse simulation behaviour.

The workbench provides:

1. A **visual graph editor** for constructing SNN topologies as directed graphs with configurable synaptic weights, input spike generators (periodic, Poisson, burst, custom), and multi-generator combination modes (OR, AND, XOR).
2. An **isomorphic simulation engine** that executes the LIF dynamics with configurable model complexity: three presets—Deterministic (1 threshold level, no refractory), Fast (4 levels, no refractory), and Full (10 levels, ARP/RRP enabled)—allow trading biological fidelity for computational tractability. Results are visualized via raster plots, membrane potential traces, and aggregate firing statistics.
3. **Automated PRISM code generation** that translates the SNN graph into a DTMC model. The generator supports both *precise* and *weight-discretized quotient* abstraction modes (Section 4), and synthesizes PCTL properties for reachability, safety, and liveness verification. Per-neuron potential bounds are computed from *fan-in analysis*—bounding each neuron's reachable potential from the signs and magnitudes of its incoming weights—to minimize the state space.
4. A **verification bridge** that invokes the PRISM model checker as a background process with configurable engine selection (explicit, sparse, MTBDD), JVM memory limits, and solver options (Gauss–Seidel, topological value iteration). Results are parsed and displayed inline alongside the simulation output.

Since model complexity presets (Deterministic, Fast, Full) are shared between simulation and verification, any configuration tested in simulation can be directly verified, enabling a rapid *design–simulate–verify* workflow without manually constructing PRISM models.

## 6 Topology-Dependent Scaling Limits

The weight-discretized quotient abstraction promises exponential state space reduction; this section quantifies that promise. We derive closed-form formulas for the DTMC state space as a function of network topology and model configuration, and validate them empirically to identify the practical limits of PRISM-based SNN verification.

### 6.1 PRISM Module Decomposition

The PRISM model for an SNN $G = (V, E)$ consists of four module types:

- **GlobalClock**: a step counter with $T_{\max}+1$ states, where $T_{\max}$ is the verification horizon.
- **Inputs**: $2^{|V_{\text{in}}|}$ states (one binary variable per input neuron).
- **Neuron** $M_n$: per-neuron state depends on configuration:
  - *Fast* ($k=10$ threshold levels, no refractory): $|S_n^{\text{fast}}| = 2\cdot R_n$ where $R_n = P_{\max,n} - P_{\min,n} + 1$.
  - *Full* ($k=10$, ARP=2, RRP=4): $|S_n^{\text{full}}| = 3\cdot 3\cdot 5\cdot 2\cdot R_n = 90\cdot R_n$.
  - *Discretized* ($W=3$): $|S_n^{\text{disc}}| = 2\cdot\left(P_{\max,n}^d + 1\right)$ where $P_{\max,n}^d = T_d + E_n^d$ and $E_n^d = \sum_{i:w_{i,n}>0}\delta_W\left(w_{i,n}\right)$ is the summed positive discretized in-weight (excitatory headroom).
- **Transfer** $T_{i,j}$: 2 states per internal edge (representing the intermediate state of a spike traveling along a synapse).

Throughout this section, *Fast* and *Full* denote precise-model refractory settings (no refractory vs. ARP/RRP) at the full $k=10$ resolution, as distinct from the reduced-level GUI presets of Section 5.

**Proposition 3**. ***(State Space Product).*** *The theoretical state space is the Cartesian product of all module state spaces:*

$$|S_{\text{theory}}| = (T_{\max}+1)\cdot 2^{|V_{\text{in}}|}\cdot\prod_{n\in V_{\text{proc}}} f_n(C)\cdot 2^{|E_{\text{int}}|} \tag{7}$$

*where* $f_n(C) = |S_n|$ *depends on the model configuration* $C$. *The* $2^{|E_{\text{int}}|}$ *factor accounts for the binary states of all internal transfer edges* $E_{\text{int}}\subseteq E$ *(synapses whose source is not an input neuron).*

For a chain of $N$ neurons with threshold $T=100$ and weight $w=80$: the per-neuron factor is $f_n = 2\cdot 121 = 242$ (fast precise), dropping to $f_n = 2\cdot 7 = 14$ (discretized $W=3$). The per-neuron reduction factor is $242/14 \approx 17.3\,\times$.

**Proposition 4**. ***(Exponential State Space Reduction).*** *For a chain of* $N$ *neurons, the state space ratio between precise and discretized models compounds exponentially:*

$$\frac{|S^{\text{precise}}|}{|S^{\text{disc}}|} = \prod_{n=1}^{N}\frac{R_n}{P_{\max,n}^d+1} \approx 17.3^N \tag{8}$$

*The factor* $17.3$ *is the per-neuron reduction ratio* $242/14$ *from the chain example above. For* $N=4$*:* $17.3^4 \approx 89,500\,\times$.

### 6.2 Empirical Validation

63 PRISM models were generated across 7 canonical topologies (single, chain-2 through chain-4, fork, diamond, convergent), 3 configurations (Det: deterministic, Fast: no refractory period, Full: complete refractory dynamics), and 3 model types (precise, disc. $W = 2$, disc. $W = 3$). Table 1 reports the reachable state counts from DTMC exports.

**Table 1.** Reachable DTMC states $|S_{\text{reachable}}|$ by topology and configuration. All precise models use $k = 10$ threshold levels. "OOM" denotes CUDD BDD out-of-memory (2 GB limit).

| Topology | Det. | Fast | Full | Disc. $W$=3 |
|---|---|---|---|---|
| Single (1N) | 5 | 6 | 18 | 6 |
| Chain-2 (2N) | 12 | 23 | 246 | 11 |
| Chain-3 (3N) | 23 | 275 | 13,180 | 28 |
| Chain-4 (4N) | 42 | 6,917 | 362,289 | 88 |
| Fork (3N) | 12 | 83 | 3,388 | 15 |
| Diamond (5N) | 15 | 1,511 | OOM | 98 |
| Convergent (3N) | 17 | 20 | 65 | 16 |

The most striking pattern is the exponential growth across chain lengths in the precise configurations: from 18 states (single, full) to 362,289 (chain-4, full)—a 20{,}000 × increase for just three additional neurons. The discretized column remains remarkably stable, growing from 6 to 88 across the same range.

The per-neuron multiplicative factor for the fast precise model is:

$$r_{\text{fast}} \approx 23/6 \approx 3.8, \quad 275/23 \approx 12.0, \quad 6917/275 \approx 25.2 \tag{9}$$

This acceleration occurs because downstream neurons receive progressively richer input distributions, approaching the theoretical maximum of 484 per neuron as the reachability density $\rho$ increases.

Three further observations emerge: (i) *fan-out is more expensive than fan-in* —the fork topology (3 neurons, fan-out 2) produces 275 fast states versus 20 for the convergent topology (fan-in 2), because parallel branches multiply module counts; (ii) *BDD memory is the practical bottleneck*—the diamond topology (5 neurons) causes CUDD OOM in the full precise configuration, while the explicit engine handles chain-4 (362,289 states); (iii) *discretization scales gracefully*—the disc. $W = 3$ diamond completes with 98 states versus OOM for the full precise model. Table 2 summarizes the estimated maximum verifiable network sizes.

**Table 2.** Estimated maximum verifiable network sizes (2 GB CUDD limit).

| Configuration | Chain | Diamond | Notes |
|---|---|---|---|
| Det. precise | >10N | >10N | Minimal state space |
| Fast precise | 5–6N | 4N | Per-neuron narrowing |
| Full precise | 4N | <4N | ARP/RRP dominate |
| Fast disc. $W$=3 | >10N | 7–8N | 17× reduction/neuron |
| Full disc. $W$=3 | 8N | 5–6N | Best precision/tractability |

## 7 Conclusion

Formal verification of spiking neural networks requires balancing the biological fidelity of probabilistic models against the compact state spaces demanded by model checking. This paper presented a weight-discretized quotient model abstraction that resolves this tension. The discretization function $\delta_W$ maps continuous synaptic weights to a compact integer range while bounding any firing disagreement to a quantization gray zone (Theorem 1) and preventing spurious sustained firing (Theorem 2). The core biological properties of LIF neurons —tonic spiking, integrator behaviour, and excitability [13] —are maintained under the multiplicative-leak update $p'_n = \lfloor r \cdot p_n \rfloor + C_n$. These guarantees are delivered through CogSpike, the unified design–simulate–verify workbench of Section 5.

The topology-dependent scaling analysis demonstrates that the state space reduction compounds exponentially across neurons: approximately 17 × per neuron for $W = 3$, enabling verification of networks that are otherwise intractable. Empirical validation across seven canonical topologies confirms the theoretical predictions and identifies BDD memory as the binding practical constraint.

Future directions include *compositional verification* exploiting neuronal archetypes [14], *automated $W$ selection* based on fan-in analysis, extension to *recurrent topologies*, and integrating verification into biologically plausible learning frameworks [18] as modulatory safety constraints.

## Bibliography

1. Baier, C., Katoen, J.-P.: Principles of Model Checking. MIT Press (2008).
2. Maass, W.: Networks of Spiking Neurons: The Third Generation of Neural Network Models. Neural Networks. 10, 1659–1671 (1997).

3. Hodgkin, A.L., Huxley, A.F.: A Quantitative Description of Membrane Current and Its Application to Conduction and Excitation in Nerve. The Journal of Physiology. 117, 500–544 (1952).

4. Izhikevich, E.M.: Simple model of spiking neurons. IEEE Transactions on neural networks. 14, 1569–1572 (2003).

5. Nguyen, D.-A., Tran, X.-T., Iacopi, F.: A Review of Algorithms and Hardware Implementations for Spiking Neural Networks. Journal of Low Power Electronics and Applications. 11, 23 (2021).

6. Kwiatkowska, M.Z., Norman, G., Parker, D.: PRISM 4.0: Verification of Probabilistic Real-Time Systems. In: Computer Aided Verification - 23rd International Conference, CAV 2011, Snowbird, UT, USA, July 14-20, 2011. Proceedings. pp. 585–591 (2011). https://doi.org/10.1007/978-3-642-22110-1_47.

7. Gerstner, W., Kistler, W.M.: Spiking Neuron Models: Single Neurons, Populations, Plasticity. Cambridge University Press (2002).

8. Brunel, N., Hakim, V.: Fast Global Oscillations in Networks of Integrate-and-Fire Neurons with Low Firing Rates. Neural Computation. 11, 1621–1671 (1999).

9. Stimberg, M., Brette, R., Goodman, D.F.M.: Brian 2, an Intuitive and Efficient Neural Simulator. eLife. 8, e47314 (2019).

10. Bekolay, T., Bergstra, J., Hunsberger, E., DeWolf, T., Stewart, T.C., Rasmussen, D., Choo, X., Voelker, A., Eliasmith, C.: Nengo: A Python Tool for Building Large-Scale Functional Brain Models. Frontiers in Neuroinformatics. 7, 48 (2014).

11. Gewaltig, M.-O., Diesmann, M.: NEST (NEural Simulation Tool). Scholarpedia. 2, 1430 (2007).

12. Hazan, H., Saunders, D.J., Khan, H., Patel, D., Sanghavi, D.T., Siegelmann, H.T., Kozma, R.: BindsNET: A Machine Learning-Oriented Spiking Neural Networks Library in Python. Frontiers in Neuroinformatics. 12, 89 (2018). https://doi.org/10.3389/fninf.2018.00089.

13. De Maria, E., Di Giusto, C., Laversa, L.: Spiking neural networks modelled as timed automata: with parameter learning. Nat. Comput. 19, 135–155 (2020). https://doi.org/10.1007/S11047-019-09727-9.

14. De Maria, E., Bahrami, A., L'Yvonnet, T., Felty, A., Gaffé, D., Ressouche, A., Grammont, F.: On the Use of Formal Methods to Model and Verify Neuronal Archetypes. Frontiers of Computer Science. 16, 163404 (2022). https://doi.org/10.1007/S11704-020-0029-6.

15. Yao, Z., De Maria, E., Simone, R. de: Probabilistic Spiking Neural Networks: Formal Verification and Simulation. In: Formenti, E. and Manzoni, L. (eds.) Unconventional Computation and Natural Computation - 22nd International Conference, UCNC 2025, Nice, France, September 1-5, 2025, Proceedings. pp. 100–114. Springer (2025). https://doi.org/10.1007/978-3-032-15641-9_8.

16. Hansson, H., Jonsson, B.: A logic for reasoning about time and reliability. Formal aspects of computing. 6, 512–535 (1994).

17. Katoen, J.-P., Kemna, T., Zapreev, I.S., Jansen, D.N.: Bisimulation Minimisation Mostly Speeds Up Probabilistic Model Checking. In: Tools and Algorithms for the Construction and Analysis of Systems, 13th International Conference, TACAS 2007. pp. 87–101 (2007). https://doi.org/10.1007/978-3-540-71209-1_9.

18. Bellec, G., Scherr, F., Hajek, A., Salaj, D., Legenstein, R., Maass, W.: Eligibility Traces Provide a Data-Inspired Alternative to Backpropagation Through Time. In: Real Neurons Hidden Units: Future Directions at the Intersection of Neuroscience and Artificial Intelligence @ NeurIPS (2019).